# From the EM Algorithm to the CM-EM Algorithm for Global Convergence of Mixture Models

Chenguang Lu
lcguang@foxmail.com

**Abstract:** The Expectation-Maximization (EM) algorithm for mixture models often results in slow or invalid convergence. The popular convergence proof affirms that the likelihood increases with *Q*; *Q* is increasing in the M-step and non-decreasing in the E-step. The author found that (1) *Q* may and should decrease in some E-steps; (2) The Shannon channel from the E-step is improper and hence the expectation is improper. The author proposed the CM-EM algorithm (CM means Channel's Matching), which adds a step to optimize the mixture ratios for the proper Shannon channel and maximizes G, average log-normalized-likelihood, in the M-step. Neal and Hinton's Maximization-Maximization (MM) algorithm use *F* instead of *Q* to speed the convergence. Maximizing *G* is similar to maximizing *F*. The new convergence proof is similar to Beal's proof with the variational method. It first proves that the minimum relative entropy equals the minimum *R-G* (*R* is mutual information), then uses variational and iterative methods that Shannon et al. use for rate-distortion functions to prove the global convergence. Some examples show that *Q* and *F* should and may decrease in some E-steps. For the same example, the EM, MM, and CM-EM algorithms need about 36, 18, and 9 iterations respectively.

**Graphical Abstract:**

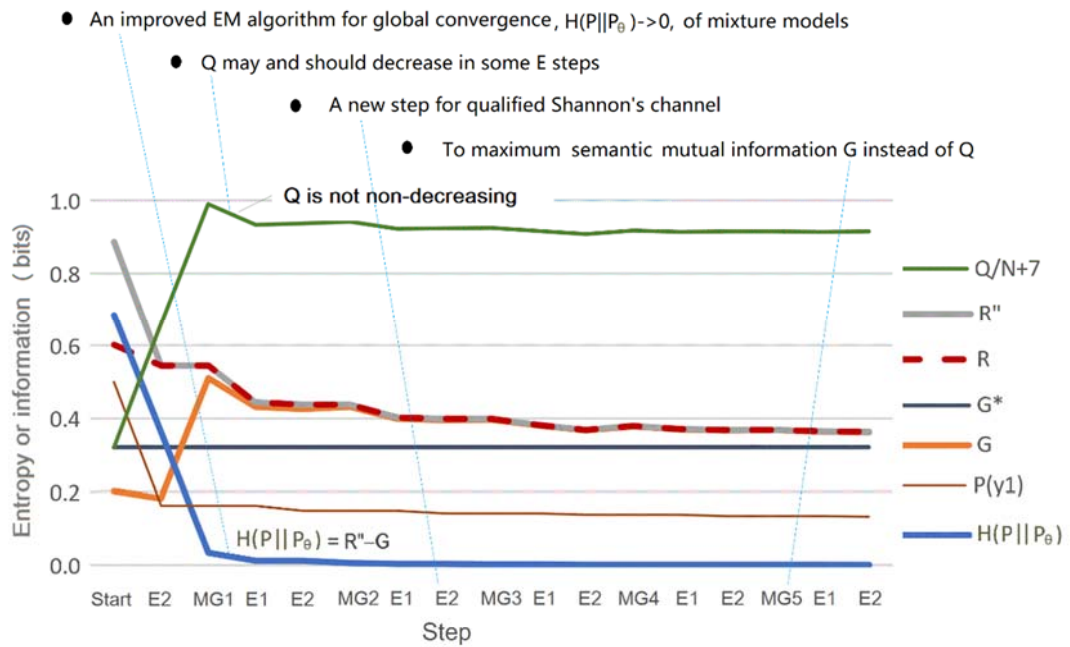

## 1. Introduction

The expectation-maximization (EM) algorithm is an iterative method, in which the model depends on unobserved latent variables. It is usually used for hidden Markov models and mixture models. This paper only discusses the EM algorithm for mixture models. The EM algorithm was explicitly proposed by Dempster, Laird, and Rubin (1977). Although there are many successful examples with the EM algorithm, this algorithm for mixture models often results in slow or invalid convergence especially when the overlap is large (Ma, Xu, and Jordan, 2000), start parameters are improper (Karlis and Evdokia, 2003; Yu, Chaomu, and Yang, 2018), or mixture ratios are unbalanced (Naim and Gildea, 2012). So, there are many improved EM algorithms (see Meng and Dyk (1997) and Roche (2012)). A significantly improved EM algorithm proposed by Neal and Hinton (1999) may be called the maximization-maximization (MM) algorithm. New algorithm in this paper has an M-step similar to that in the MM algorithm.

To prove the convergence of the EM algorithm, in the famous paper of Dempster *et al.* and the papers improving the convergence proof (Wu, 1983; Little and Rubin, 1987), these authors affirm:

**Affirmation I**: We may achieve the maximum likelihood by increasing $Q$ repeatedly.

**Affirmation II**: $Q$ is increasing in the M-step and is non-decreasing in the E-step.

The above two affirmations to hidden Markov models should be correct. However, to mixture models, they should be wrong. In many papers and webpages introducing the EM algorithm, the two affirmations still exist. For example, on website Wikipedia, under topic "Expectation–maximization algorithm", there is a section with the title "Proof of correctness". we can see "Expectation-maximization works to improve $Q(\theta|\theta^{(t)})$ rather than directly improving log$P(\mathbf{X}|\theta)$. Here is shown that improvements to the former imply improvements to the latter." [1] We call such proof the popular convergence proof.

There are also some researchers, such as Beal (2003) and Davison (2003), who provided the convergence proofs that do not require that $Q$ is non-decreasing in the E-step. Beal's proof is followed by this paper. However, most scholars still believe the popular convergence proof so that we hardly see such a statement that the E-step may decrease $Q$.

The author of this paper argues that 1) the EM algorithm for mixture models is imperfect in theory; 2) the popular convergence proof for mixture models is incorrect; 3) before improving the EM algorithm, we should reconsider the significance and limitation of the E-step. The author found two reasons:

**The first reason**: In some cases, the E-step may and should decrease $Q$ (see Section 2.3 for details).

**The second reason**: The conditional distribution or Shannon's channel $P(Y|X)$ from the E-step is improper for calculating the expectation because that $\sum_i P(x_i)P(y_j|x_i) \neq P(y_j)$ and hence $P(X|y_j)$ is not normalized in general (see Section 2.4 for details).

The author also found:
- The EM algorithm can converge in most cases not because the E-step is non-decreasing but because the E-step decreases the relative entropy or Kullback-Leibler divergence (see Eq. (2.6)). That means that the mistake in Affirmation II covers up the mistake in Affirmation I.
- Although the MM algorithm can speed the convergence of mixture models, $F$ should also decrease in some E-steps.
- The EM algorithm lacks a step for modifying mixture ratio vector $P(Y)$ so that $P(Y|X)$ from the E-stem is improper. There is an improved EM or MM algorithm, the CM-EM algorithm, which can make mixture models converge much better than the EM algorithm.

However, the basic idea behind the CM-EM algorithm is different from that behind the EM algorithm. The CM means Channels' Matching, e. g., that the semantic channel and the Shannon channel mutual match. In the CM-EM algorithm, the M-step maximizes the semantic mutual information $G$ whereas the E-step minimizes Shannon's mutual information $R$. This algorithm can minimize $R$-$G$ and hence can minimize the relative entropy in every step.

---

[1] https://en.wikipedia.org/wiki/Expectation%E2%80%93maximization_algorithm

The CM-EM algorithm is also one kind of the CM algorithm. There is also a CM algorithm for maximum mutual information classifications (Lu, 2018a), where the step for the Shannon channel is no longer similar to the E-step of the EM algorithm; it also maximizes *R* so that both *R* and *G* reach their maxima (see Section 3.4 or Lu, 2018a).

In the recent two decades, the cross-entropy method has been used for statistical learning and played an important role (Goodfellow and Bengio, 2016). The author extended Shannon's information theory (Shannon, 1948) into a generalized information theory twenty years ago (Lu, 1994 and 1999). The generalized mutual information is a mutual cross-entropy. It is also called semantic mutual information. In this theory, a set of truth functions form a semantic channel, the truth function and the likelihood function can be mutually converted from one to another, and the semantic mutual information is defined with average log(normalized-likelihood).

Recently, the author (Lu, 2017; Lu, 2018a) found the CM algorithm. However, in that paper on the CM algorithm for mixture models (Lu, 2017), e. g., the CM-EM algorithm, the convergence proof is not strict, and the relationships between the CM-EM algorithm and the EM and MM algorithms are not clear enough. Now we use the cross-entropy method to clarify the above relationships and use the variational method and iterative method used by Shannon (1959) *et al.* (Berger, 1971; Zhou, 1983) for analyzing *R*(*D*) function to strictly prove that the CM-EM algorithm can globally converge. It should be mentioned that Beal (2003) used a similar variational method earlier to prove that the E-step of the EM algorithm for mixture models increases the likelihood log*P*(**X**|*θ*), instead of *Q*. The difference is that the convergence proof of the CM-EM algorithm also needs to prove that the new step for optimizing mixture ratio vector *P*(*Y*) can increase the likelihood or decrease the relative entropy.

The rest of this paper is organized as follows. Section 2 uses cross-entropies as tools to explain the EM algorithm for mixture models and to reveal the problems with the EM algorithm. Section 3 introduces the CM-EM algorithm, including the convergence proof, iterative examples, and the maximum mutual information classification. Section 4 clarifies the relationship between the EM, MM, and CM-EM algorithms and compares the numbers of iterations of using these algorithms. Section 5 contains discussion. Section 6 ends with conclusions.

The package with Excel files and Word files illustrating the CM algorithm for mixture models and maximum mutual information classifications can be obtained from http://survivor99.com/lcg/CM-iteration.zip. These Excel files also contain the data of iterative processes.

## 2. Using Cross-entropies to Explain Problems with the EM Algorithm for Mixture Models

### 2.1 Mathematical Definitions

**Definition 1** Let *U* be an instance space and *X* be a random variable taking value from *U*={$x_1, x_2, ..., x_m$}. Let *V* be a label space and *Y* be a random variable taking value from *V*={$y_1, y_2, ..., y_n$}. For analytical convenience, we assume that both *U* and *V* are one dimensional.

**Definition 2** A sample **D** consists of a number of examples, e. g. **D**={(*x*(*t*); *y*(*t*)|*t*=1, 2, ..., *N*; *x*(*t*) ∈ *U*; *y*(*t*) ∈ *V*}, where *t* is the sequence number and *N* is the number of examples. A conditional sample is **D**$_j$={(*x*(*t*); $y_j$)|*t*=1, 2, ..., $N_j$; *x*(*t*) ∈ *U* }, where $N_j$ is the number of examples with $y_j$.

**Definition 3** Let *θ* be a predictive model. For each $y_j$, there is a sub-model $θ_j$ and a predictive distribution *P*(*X*|$θ_j$), which is also the likelihood function of $θ_j$. Hence the predicted distribution of *X* is

$$P_\theta(X) = \sum_j P(y_j) P(X | \theta_j). \qquad (2.1)$$

We use *P*(*X*|$θ_j$) instead of *P*(*X*|$y_j$, *θ*) as in popular methods because using *P*(*X*|$θ_j$), we can clearly show relationships between various cross-entropies and corresponding Shannon's entropies.

**Definition 4** Let $P(X)$ be an instance distribution, $P(Y)$ be a label distribution, and $P(X|y_j)$ be a conditional instance distribution, from statistics of **D** or **D**$_j$. The cross-entropy of $P_\theta(X)$ relative to $P(X)$ is

$$H_\theta(X) = -\sum_i P(x_i) \log P_\theta(x_i).  \quad (2.2)$$

The cross-entropy of $P(X|\theta_j)$ relative to $P(X|y_j)$ is

$$H(X|\theta_j) = -\sum_i P(x_i|y_j) \log P(x_i|\theta_j).  \quad (2.3)$$

Akaike (1974) proved that the maximum likelihood criterion is equivalent to the minimum Kullback-Leibler (KL) divergence (1951) criterion. The KL divergence is equal to the cross-entropy minus the Shannon entropy, e. g., $H(P||P_\theta)=H_\theta(X)-H(X)$. Following Akaike, we prove that the average log likelihood is equal to the negative cross-entropy as follows.

**Proof**: In Definition 2, if instances in **D**$_j$ are produced by $N_j$ ($N_j\to\infty$) Independent and Identically Distributed (IID) random variables (so assumed hereafter). Suppose the number of $x_i$ in **D**$_j$ is $N_{ij}$. When $N_j\to\infty$, $P(x_i|y_j)= N_{ij}/N_j$. With the IID assumption, the log likelihood of $\theta_j$ is

$$\log L_X(\theta_j) = \log P(x(1),x(2),...,x(N_j)|\theta_j) = \log \prod_i P(x_i|\theta_j)^{N_{ji}}$$
$$= N_j \sum_i P(x_i|y_j) \log P(x_i|\theta_j) = -N_j H(X|\theta_j). \quad (2.4)$$

Hence, its average is $L_X(\theta_j)/N_j=-H(X|\theta_j)$. **QED.**

It is easy to prove that when $P(X|\theta_j)=P(X|y_j)$, for all $j$, the cross-entropy or the likelihood reaches its maximum and the KL divergence reaches its minimum.

Similarly, the likelihood of $\theta$ for given **D** is $\log L_X(\theta)= -NH_\theta(X)$. $L_X(\theta)$ is $P(\mathbf{X}|\theta)$ above. The objective function $Q$ in the EM algorithm is $\log L_{X,Y}(\theta)$ or $\log(\mathbf{X},\mathbf{Y}|\theta)$. The relationship between $Q$ and the joint cross-entropy is

$$Q = \log L_{X,Y}(\theta) = \log \prod_j [P(y_j)^{N_j} P(x(1),x(2),...,x(N_j)|\theta_j)]$$
$$= N\sum_i P(x_i,y_j) \log P(x_i,y_j|\theta_j)=-NH(X,Y|\theta). \quad (2.5)$$

Thus, $Q$ can be treated as a negative joint cross-entropy $-H(X,Y|\theta)$.

## 2.2 Using Cross-entropies to Explain the EM algorithm for Mixture Models

Assume that $n$ Gaussian distribution functions (true models) are :

$$P^*(X|y_j) =P(X|\theta_j^*)=K_j\exp[-(X-\mu_{j*})^2/(2\sigma_j^{*2})], \ j=1, 2, …, n,$$

where $K_j$ is a normalizing coefficient, $\mu_j$ is the mean, and $\sigma_j$ is the standard deviation. In the following, we assume $n=2$. Then the instance distribution $P(X)$ is the mixture of two Gaussian distributions:

$$P(X)=P^*(y_1)P^*(X|y_1)+P^*(y_2)P^*(X|y_2),$$

where $P^*(y_1)$ and $P^*(y_2)=1-P^*(y_1)$ are two true mixture ratios. We only know $P(X)$ and $n=2$ without knowing true model parameters $\mu_1^*, \mu_2^*, \sigma_1^*, \sigma_2^*$ and the true mixture ratios. We can guess the distribution:

$$P_\theta(X)=P(y_1)P(X|\theta_1)+P(y_2)P(X|\theta_2).$$

The relative entropy or KL divergence of $P_\theta(X)$ relative to $P(X)$ is

$$H(P||P_\theta) = \sum_i P(x_i) \log \frac{P(x_i)}{P_\theta(x_i)} = H_\theta(X) - H(X). \quad (2.6)$$

If two distributions are close to each other so that the relative entropy is close to 0, say, less than 0.001 bit, then we may say that our guess is right. Therefore, our task is to change $P(Y)$ and $\theta$ to maximize likelihood $L_X(\theta)$ or to minimize relative entropy $H(P||P_\theta)$.

The main formula of the EM algorithm for mixture models (Dempster *et al.*, 1977) can be described as follows:

$$\log L_X(\theta) = N \sum_i P(x_i) \log P_\theta(x_i) = N \sum_i P(x_i) \log \sum_j P(x_i|\theta_j) P(y_j)$$

$$\geq L = N \sum_i \sum_j P(x_i) P(y_i|x_i) \log \frac{P(x_i, y_j|\theta)}{P(y_i|x_i)} \quad (2.7)$$

$$= -NH(X,Y|\theta) + NH(Y|X,\theta) = Q(\theta|\theta^{(t)}) - H$$

where $\theta^{(t)}$ means $\theta$ before M-step. The inequality sign is used because of Jensen's inequality. Let $Q(\theta|\theta^{(t)})$ be simply denoted by $Q$. There is $Q = L_{X,Y}(\theta) = -NH(X, Y|\theta)$. Hence, to optimize a mixture model is to maximize $L = Q - H$.

Steps in the EM algorithm are:

**E-step:** Write the conditional probability functions (e. g., the Shannon channel):

$$P(y_j|X) = P(y_j)P(X|\theta_j)/P_\theta(X), \ j=1,2,...,n;$$
$$P_\theta(X) = \sum_j P(y_j)P(X|\theta_j). \quad (2.8)$$

**M-step**: Improve $P(Y)$ and $\theta$ to maximize $Q$. If $Q$ cannot be improved further, then end the iteration; otherwise, go to the E-step.

Dempster *et al.* (1977), Wu (1983) and Little *et al.* (1985) affirm that 1) We can maximize $L_X(\theta)$ by increasing $Q$; 2) The M-step can increase $Q$, and the E-step does not decrease $Q$. If so, the iteration can converge by repeating the M-step and E-step. However, it was never well proved that $Q$ is non-decreasing in the E-step!

In the MM algorithm (Neal and Hinton, 1999), an improved EM algorithm, the objective function $Q$ is changed into $F = Q + NH(Y)$, where $H(Y)$ is the Shannon entropy of $Y$. The MM algorithm maximizes $F$ in both M-step and E-step so that the convergence is faster than that of the EM algorithm. However, it was never proved that $F$ and $L_X(\theta)$ are positively related. We can also find counterexamples (see Example 2 in Section 3.3).

### 2.3 A Counterexample Where Q May and Should Decrease

The true mixture ratio vector $P^*(Y)$ and the true conditional probability distributions $P^*(X|Y)$ ascertain the joint probability distribution $P^*(X, Y)$. The corresponding joint entropy is

$$H^*(X,Y) = -\sum_j \sum_i P^*(x_i|y_j^*)P^*(y_j)\log[P^*(x_i|y_j)P^*(y_j)] = -Q^*/N. \quad (2.9)$$

According to Eq. (2.5), the joint cross-entropy is $H(X, Y|\theta) = -Q/N$. Now we show that $H(X, Y|\theta)$ may be less than $H^*(X, Y)$ and hence $Q$ may be larger than $Q^*$.

$H^*(X,Y) = H^*(Y|X) + H^*(Y)$ has a certain value. Assume $n=2$ and $P(y_1)=P(y_2)=0.5$. Then $H^*(Y)=1$ bit. The bigger the $\sigma_1^*$ and $\sigma_2^*$ are, the bigger the $H^*(X|Y)$ is. If $H^*(X, Y)$ is very large whereas $H(X, Y|\theta)$ is not big enough, then $Q$ will be bigger than $Q^*$. In this case, if we still increase $Q$, we must go in the wrong direction.

For example, $U = \{1, 2, 3, …, 100\}$, true model ratio $P^*(y_1)=0.5$, true model parameters $\mu_1^*=35$, $\mu_2^*=65$, and $\sigma_1^* = \sigma_2^* = 15$. Assume that the guessed ratios and parameters are $P(y_1) = P(y_2) = 0.5$, $\mu_1 = \mu_1^*$, $\mu_2 = \mu_2^*$, and $\sigma_1 = \sigma_2 = \sigma$. Table 1 shows how the left $\sigma$ and right $\sigma$ (new $\sigma$) make $Q > Q^*$ (The left/right $\sigma$ is the $\sigma$ on the left/right of the log in Eq. (2.5)).

**Table 1.** Counterexamples against the Popular Convergence Proof of the EM Algorithm

|          | left $\sigma$ | right $\sigma$ | $Q/N = -H(X,Y|\theta)$(bits) |
|----------|---------------|----------------|------------------------------|
| True $Q^*$ | 15 | 15 | -6.89 |
| Larger $Q$ | 10 | 10 | -6.75, counterexample! |
| Larger $Q$ | 5  | 12 | -6.59, counterexample! |

$H^*(X, Y)$ is 6.89 bits, and hence $Q^*$ is -6.89$N$ bits, which is rather small. If start parameters $\sigma_1=\sigma_2=10$ and $P(y_1)=P^*(y_1)=0.5$, then $Q=-6.75N$ bits$>Q^*$. If the left $\sigma=5$ and the right $\sigma$ is 12, then $H(X, Y|\theta)$ is still smaller, and hence $Q$ is still larger than $Q^*$. In these cases, if we still increase $Q$, we will go in the wrong direction.

Now we consider the second mistake in the popular convergence proof of the EM algorithm. The authors affirm that the E-step never decreases $Q$. However, there also exist counterexamples. Fig. 3 for Example 2 in Section 3.3 shows that the E-step, e. g., E1-step of the CM-EM algorithm, may decrease $Q$. When $H(X,Y|\theta)<H^*(X, Y)$, it is because the E-step may increase $H(X, Y|\theta)$ or decrease $Q$, $H(X, Y|\theta)$ may converge to $H^*(X, Y)$ and hence $Q$ may converge to $Q^*$. Therefore, in the popular convergence proof, the second mistake covers up the first mistake.

The MM algorithm (Neal and Hinton, 1999) uses $F=Q+NH(Y)$ as the objective function. Although the MM algorithm can speed the iterative convergence, it has a similar problem because it also maximizes $F$ in the E-step as in the M-step. However, in some cases, $F$ should also be decreased in E-steps for the convergence (see Example 2 in Section 3.3).

*2.4 The Problem with the Shannon Channel P(Y|X) from the E-step*

From the observed data, we obtain $P(X)$. From Eq. (2.8), we obtain $P(Y|X)$.
For given $P(X)$, $P(Y)$ and $\theta$, we have new $P(Y)$ denoted by $P^{+1}(Y)$:

$$P^{+1}(y_j) = \sum_i P(x_i)P(y_j|x_i) = \sum_i P(x_i)P(x_i|\theta_j)P(y_j)/P_\theta(x_i), j=1,2,...n \quad (2.9)$$

Generally, $P^{+1}(Y) \neq P(Y)$. From $P(X)$, $P(Y|X)$, and $P(Y)$, we can obtain $P(X|Y)$:

$$P(X|y_j) = P(X)P(X|\theta_j)/P_\theta(X), j=1,2,...,n$$

Generally, this $P(X|y_j)$ is not normalized. In Example 2 of Section 3.2, after the start step, $P(X|y_j)$ is so ridiculous that $\sum_i P(x_i|y_1)<0.4$ and $\sum_i P(x_i|y_2)>1.6$.

So, for given $P(X)$ and $\theta$, we need to find the unique $P(Y)$ that matches $P(X)$ and $\theta$ so that $P^{+1}(Y)=P(Y)$ and $P(X|y_j)$ is normalized.

## 3. The CM-EM Algorithm and the Convergence Proof

*3.1 The CM-EM Algorithm*

Shannon calls $P(y_j|X)$ with certain $y_j$ and variable $X$ the transition probability function from $X$ to $y_j$ (1948). A set of transition probability functions form a Shannon's channel: $P(y_j|X), j=1, 2, ..., n$.
Using $P(y_j|X)$, we can make Bayes' prediction

$$P(X|y_j) = P(X)P(y_j|X)/P(y_j), \ P(y_j) = \sum_i P(x_i)P(y_j|x_i). \quad (3.1)$$

**Definition 5.** Assume that $\theta_j$ is also a fuzzy set of $U$, $T(\theta_j|X)$ is its membership function (Zadeh, 1965), and $y_j$="X is in $\theta_j$." Then $T(\theta_j|X)$ is also the truth function of $y_j$. A set of truth value functions or membership functions form a semantic channel: $T(\theta_j|X), j=1, 2, ..., n$.

The truth function $T(\theta_j|X)$ can also be used for Bayes' prediction to produce the likelihood function:

$$P(X|\theta_j) = P(X)T(\theta_j|X)/T(\theta_j), \; T(\theta_j) = \sum_i P(x_i)T(\theta_j|x_i). \qquad (3.2)$$

Letting the maximum of $T(\theta_j|X)$ be 1, we have $T(\theta_j)=1/\max[P(X|\theta_j)/P(X)]$ and

$$T(\theta_j|X)=P(X|\theta_j)/P(X)/\max[P(X|\theta_j)/P(X)]. \qquad (3.3)$$

The author proposed the Third Kind of Bases' Theorem for logical Bayesian inference (Lu, 2018b), which includes Eqs. (3.2) and (3.3).

The semantic mutual information is defined with average log-normalized-likelihood (Lu, 1993):

$$I(X;\theta) = \sum_j \sum_i P(x_i)P(y_j|x_i)\log\frac{P(x_i|\theta_j)}{P(x_i)} = \sum_j \sum_i P(x_i)P(y_j|x_i)\log\frac{T(\theta_j|x_i)}{T(\theta_j)}. \qquad (3.4)$$

The author extended Shannon's rate-distortion function $R(D)$ to rate-semantic-mutual-information function $R(G)$ for semantic communication optimization including data compression according to visual discrimination (Lu, 1999). $R(G)$ function can also be regarded as rate-average-log-normalized-likelihood function. $R(D)$ is the minimum of Shannon's mutual information $R$ for given the upper limit $D$ of average distortion whereas $R(G)$ is the minimum of $R$ for given lower limit $G$. $R(D)$ function is concave. $R(G)$ function is a natural extension of $R(D)$ function and hence is also concave. It is bowl-shaped as shown in Fig.1.

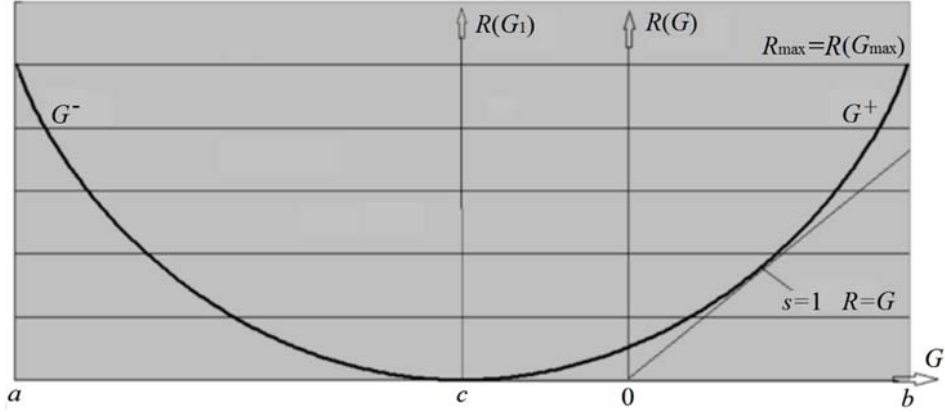

Fig. 1. Any $R(G)$ function is bowl-shaped. It has a point where $R=G$, which implies $P(X|\theta_j)=P(X|y_j)$ ($j=1,2, \ldots$).

For mixture models, we need to minimize $R-G$ so that $P(X|\theta_j)= P(X|y_j)$ ($j=1, 2, \ldots$). The CM-EM algorithm is to optimize $P(y_j|X)$ on the left of the log and $P(X|\theta_j)$ on the right of the log repeatedly. The relationships between $G$ and other several cross-entropies are:

$$\begin{aligned} G &= I(X;\theta)=H(X)-H(X|\theta) = H(X)+H_\theta(Y)-H(Y,X|\theta), \\ H(X|\theta) &= -\sum_j \sum_i P(x_i,y_j)\log P(x_i|\theta_j), \\ H_\theta(Y) &= -\sum_i P^{+1}(y_j)\log P(y_j), \end{aligned} \qquad (3.5)$$

where $H(X)$ is the Shannon entropy of $X$; $H(X|\theta)$ is the posterior entropy of $X$ and similar to $F$ in the MM algorithm; $H_\theta(Y)$ is the cross-entropy instead of the Shannon entropy; $P^{+1}(y_j)$ is the probability of $y_j$ obtained in the E2-step by

$$P^{+1}(y_j) = \sum_i P(x_i)P(y_j|x_i) = \sum_i P(x_i)P(y_j)P(X|\theta_j)/P_\theta(X). \quad (3.6)$$

Hence, there is

$$-Q/N = H(X,Y|\theta) = H(X) + H_\theta(Y) - G = H_\theta(Y) + H(X|\theta).$$

The CM-EM algorithm has three steps.

**E1-step:** Construct the Shannon channel $P(Y|X)$. This step is the same as the E-step of the EM algorithm.

**E2-step:** Modify $P(Y)$ so that $P^{+1}(Y) = P(Y)$. We may use the following tree equations repeatedly until $P^{+1}(Y)$ converges to $P(Y)$.

$$\begin{aligned}
P^{+1}(y_j) &= \sum_i P(x_i)P(y_j|x_i), j=1,2,...,n; \\
P(y_j) &= P^{+1}(y_j); j=1,2,...,n; \\
P(y_j|x_i) &= P(x_i|\theta_j)P(y_j)/\sum_k P(y_k)P(x_i|\theta_k), i=1,2,...,m; j=1,2,...,n.
\end{aligned} \quad (3.7)$$

Repeating Eq. (3.7) makes the relative entropy $H(P||P_\theta)$ less rather than makes $Q$ larger. Later, we will prove that every step in Eq. (3.7) minimizes $H(P||P_\theta)$ and hence maximizes $L_X(\theta)$ or $L(\mathbf{X}|\theta)$.

If $H(P||P_\theta)$ is less than a tiny value, such as 0.0001 bit, then end the iteration.

**MG-step:** With Eq. (3.4), fix the left part of the log and optimize the parameters of likelihood functions on the right of the log to maximize $G$. Then go to E1-step.

According to Eq. (3.4), $G$ reaches the maximum when $P(X|\theta_j^{+1})/P(X) = P(X|\theta_j)/P_\theta(X)$. Hence, the new likelihood function is

$$P(X|\theta_j^{+1}) = P(X)P(X|\theta_j)/P_\theta(X). \quad (3.8)$$

Without E2-step, $P(X|\theta_j^{+1})$ above is not normalized. If $P(X|\theta_j)$ is a Gaussian distribution, we can easily obtain new parameters:

$$\begin{aligned}
\mu_j^{+1} &= \sum_i P(x_i|\theta_j^{+1})x_i, j=1,2,...,n; \\
\sigma_j^{+1} &= \{\sum_i P(x_i|\theta_j^{+1})[x_i - \mu_j^{+1}]^2\}^{0.5}, j=1,2,...,n.
\end{aligned} \quad (3.9)$$

If the likelihood functions are not Gaussian distributions, we can find optimized parameters by searching the parameter space or using the Gradient Descent method. The author's experiment shows that for Gaussian likelihood functions, the new parameters derived from Eq. (3.9) are very similar to those obtained by optimizing $2n$ parameters successively.

### 3.2 The convergence proof of the CM-EM algorithm for mixture models

This proof makes use of the properties of $R(G)$ function (Lu, 1999):
- $R(G)$ function is concave and $R(G)-G$ has the exclusive minimum 0 as $R(G)=G$;
- $R(G)-G$ is close to relative entropy $H(P||P_\theta)$.

After E1-step (or the E-step in the EM algorithm), Shannon's mutual information $I(X; Y)$ becomes

$$R = \sum_i \sum_j P(x_i)\frac{P(x_i|\theta_j)}{P_\theta(x_i)}P(y_j)\log\frac{P(y_j|x_i)}{P^{+1}(y_j)}. \quad (3.10)$$

We define

$$R'' = \sum_i \sum_j P(x_i) \frac{P(x_i|\theta_j)}{P_\theta(x_i)} P(y_j) \log \frac{P(x_i|\theta_j)}{P_\theta(x_i)}. \tag{3.11}$$

It is easy to prove that $R''-G=H(P||P_\theta)$. Hence

$$R = \sum_i \sum_j P(x_i) \frac{P(x_i|\theta_j)}{P_\theta(x_i)} P(y_j) \log \left[ \frac{P(x_i|\theta_j)P(y_j)}{P_\theta(x_i)P^{+1}(y_j)} \right] = R'' - H(Y^{+1}||Y),$$

$$H(Y^{+1}||Y) = \sum_j P^{+1}(y_j) \log[P^{+1}(y_j)/P(y_j)]. \tag{3.12}$$

$$H(P||P_\theta) = R'' - G = R + H(Y^{+1}||Y) - G. \tag{3.13}$$

The three steps in the CM-EM seemly just right improve $R$, $H(Y^{+1}||Y)$, and $G$ respectively. However, the difficulty in the convergence proof is that when we minimize $R$ or $H(Y^{+1}||Y)$, the other two items also change. For example, when we minimize $H(Y^{+1}||Y)$, we do not know whether it is possible that $R-G$ increases too much to reduce $R''-G$.

**The Convergence Proof:** Proving that $P_\theta(X)$ converges to $P(X)$ is equivalent to proving that $H(P||P_\theta)$ converges to 0. Since E2-step makes $R=R''$ and $H(Y^{+1}||Y)=0$, we only need to prove that every step minimizes $R-G$ (see Eq. (3.13)) after the start step.

It is evident that the MG-step minimizes $R-G$ because this step maximizes $G$ without changing $R$. The left question is how to prove that E1-step and E2-step minimize $R-G$. Fortunately, we can strictly prove that by the variational method and the iterative method that Shannon (1959) and others (Berger, 1971; Zhou, 1883) used for analyzing the rate-distortion function $R(D)$. The following analysis is little different from the rate-distortion function analysis in that the distortion $d_{ij}$ becomes $I(x_i; \theta_j)=\log[P(x_i|\theta_j)/P(x_i)]$ and the parameter $s$ for $R(D)$ becomes 1.

We use the Lagrangian multiplier method to optimize $P(Y|X)$ and $P(Y)$ respectively to minimize $I(X; Y) - I(X; \theta)$. Since $P(Y|X)$ and $P(Y)$ are interdependent, we can only fix one to optimize another. To optimize $P(Y|X)$, the restrictive condition is

$$\sum_j P(y_j|x_i) = 1, \ i=1,2,...,n.$$

To optimize $P(Y)$, the restrictive condition is

$$\sum_j P(y_j) = 1.$$

The Lagrange function is therefore

$$F = I(X;Y) - I(X;\theta) + \mu_i \sum_j P(y_j|x_i) + \alpha \sum_j P(y_j).$$

To optimize $P(Y|X)$, we fix $P(y_j)$ in $F$ and then order $\partial F / \partial P(y_j|x_i) = 0$. Hence, we derive the optimized $P(Y|X)$ (see Appendix I for details):

$$P^*(y_j|x_i) = P(y_j)P(x_i|\theta_j) / \sum_k P(y_k)P(x_i|\theta_k), \ i=1, 2, ..., n; \ j=1, 2. \tag{3.14}$$

It is this formula that is used in the E-step of the EM algorithm and in the E1-step of the CM-EM algorithm. Thus, E1-step minimizes $R-G$.

To optimize $P(Y)$, we fix $P(y_j|x_i)$ in $F$ and then order $\partial F / \partial P(y_j) = 0$. Hence, we derive the optimized $P(Y)$ (see Appendix I for details):

$$P^*(y_j) = \sum_i P(x_j)P(y_j | x_i), j=1, 2, …, n. \qquad (3.15)$$

It is this formula together with Eq. (3.14) that are used in the E2-step. Thus, every step in the E2-step minimizes *R-G*.

According to Eq. (3.13), $H(P||P_\theta)$ can converge to 0 because 1) every step minimizes *R-G*, 2) the E2-step reduces $H(Y^{+1}||Y)$ to 0. Since *R(G)-G* is concave with exclusive minimum 0 as *R(G)=G* (see Fig. 1), the convergence is global. **QED**.

This convergence proof is not limited to Gaussian distribution likelihood functions. For other likelihood functions, the convergence proof is the same. So long as the maximum likelihood estimation of each $\theta_j$ is valid in every right step, the convergence should be global.

Beal (1998) used a variational method earlier to derive the same $P^*(y_j|X)$ as that in Eq. (3.14) to prove that the E-step of the EM algorithm can maximize the likelihood $L_X(\theta)$. However, since $P(Y)$ is also the function of $P(Y|X)$, it is not enough only to optimize $P(Y|X)$ without optimizing $P(Y)$.

### *3.3 The Two Examples of Using the CM-EM Algorithm*

Two examples below show the iterative processes of the CM-EM algorithm. Assume *n*=2. To check the algorithm, we directly use two true Gaussian distributions $P(X|\theta_j^*)$, *j*=1, 2, and mixture ratios $P^*(Y)$, to produce an instance distribution $P(X)$, rather than use a sample sequence to obtain $P(X)$ by statistics. When samples are huge, the results from two methods should be the same. The Shannon mutual information between *Y* and *X* for the true mixture model is $R^*=G^*=H(X)-H^*(X|Y)$.

**Example 1.** The start $P(Y)$ and $\theta$ make $R<R^*$. *G* increases in the iterative process. Relevant data are shown in Table 2. The relative entropy $H(P||P_\theta)$, e.g., *R″-G*, and various measures of information or entropy change in the iterative process as shown in Fig.2.

**Table 2.** Real and guessed model parameters and iterative results of Example 1 (*R<R*\**)

|  | Real parameters | | | Start parameters $H(P\|\|P_\theta)$=0.680 bit | | | Parameters after 5 MG-steps $H(P\|\|P_\theta)$=0.00092 bit | | |
| --- | --- | --- | --- | --- | --- | --- | --- | --- | --- |
|  | $\mu^*$ | $\sigma^*$ | $P^*(Y)$ | $\mu$ | $\sigma$ | $P(Y)$ | $\mu$ | $\sigma$ | $P(Y)$ |
| $y_1$ | 35 | 8 | 0.7 | 30 | 15 | 0.5 | 35.4 | 8.3 | 0.720 |
| $y_2$ | 65 | 12 | 0.3 | 70 | 15 | 0.5 | 65.2 | 11.4 | 0.280 |

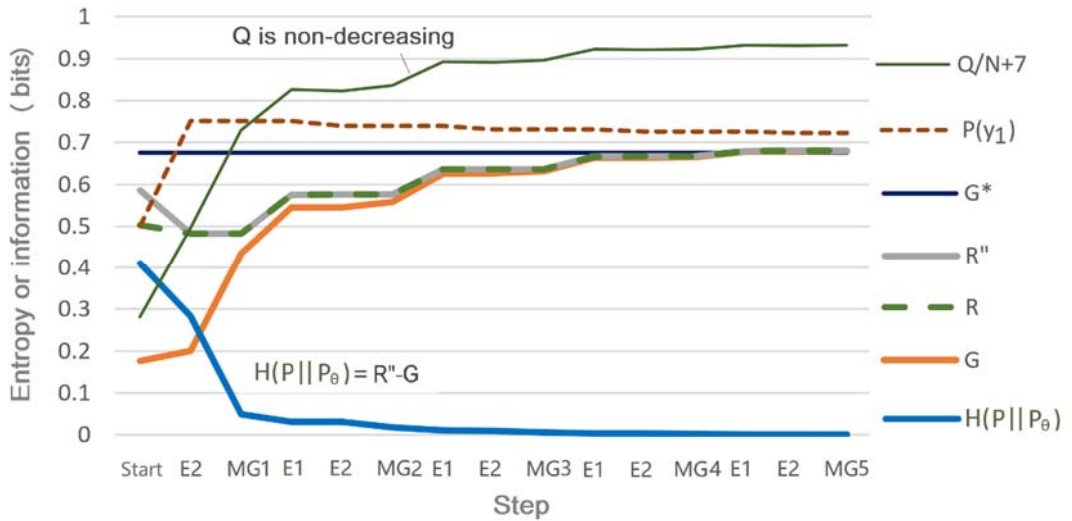

**Fig. 2.** The iterative process when *R<R\** and *G<G\**. $H(P||P_\theta)$= *R″-G* decreases in all steps. *G* and *Q* are monotonically increasing or non-decreasing.

In this example, $Q$, $G$, and $F$ are increasing or non-decreasing in every step. So, this example accords with the expectation of the authors who prove that EM or MM algorithm converges. However, the following example is against their anticipations.

**Example** 2. The start $P(Y)$ and $\theta$ make $G>G^*$. Table 3 shows relevant data. Fig. 4 shows the iterative process.

**Table 3.** Real and guessed parameters and iterative results of Example 2 ($R>R^*$)

|  | Real parameters | | | Starting parameters $H(P\|\|P_\theta)=0.68$ bit | | | Parameters after 5 MG-steps $H(P\|\|P_\theta)=0.00092$ bit | | |
|---|---|---|---|---|---|---|---|---|---|
|  | $\mu^*$ | $\sigma^*$ | $P^*(Y)$ | $\mu$ | $\sigma$ | $P(Y)$ | $\mu$ | $\sigma$ | $P(Y)$ |
| $y_1$ | 35 | 8 | 0.1 | 30 | 8 | 0.5 | 38 | 9.3 | 0.134 |
| $y_2$ | 65 | 12 | 0.9 | 70 | 8 | 0.5 | 65.8 | 11.5 | 0.866 |

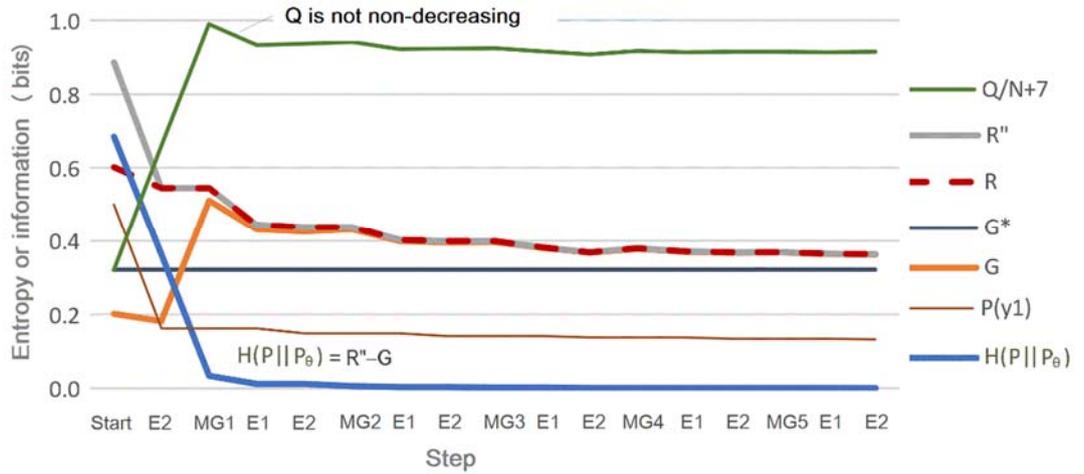

**Fig. 3.** The iterative process when $R>R^*$. $Q$ decreases in some E-steps. The line for $G$ indicates that $G$ and $F$ decrease in most E1 steps. The line for $P(y_1)$ indicates that the E2 step can optimize $P(Y)$ quickly.

In this iterative process, $Q$ decreases in the second E-step or E1-step; $G$ and $F$ (in the MM algorithm) decreases in every E1-step after the first MG-step. This example is a counterexample against both convergence proofs of the EM algorithm and the MM algorithm.

Table 4 shows the reduced relative entropy $\Delta H(P\|\|P_\theta)$ in the three kinds of steps. We can see that the E2-step plays a vital role in speeding the convergence. E2-step not only reduces $H(P\|\|P_\theta)$ but also provides optimization room for the next MG-step.

**Table 4.** The CM-EM algorithm reduces relative entropy $\Delta H(P\|\|P_\theta)$ (bits) in different steps

|  | $\Delta H(P\|\|P_\theta)$ | | | |
|---|---|---|---|---|
|  | E1-step | E2-step | M-step | Sum |
| Example 1 | 0.025 | 0.128 | 0.255 | 0.409 |
| Example 2 | 0.025 | 0.322 | 0.337 | 0.683 |

*3.4 Maximum Mutual Information Classifications for Given Mixture Models*

After we obtain the optimized mixture components $P^*(X|y_j)=P(X|\theta_j^*)$, $j=1, 2, \ldots, n$, we need to use the mixture model to classify $X$ with the corresponding unseen instance $Y$ into different classes.

The CM algorithm for maximum mutual information classifications (Lu, 2018a) can be used for this purpose.

We use $Z$ to denote the predicted class, and hence $Z$ is the function of $X$: $Z=f(X)$. We may regard $Y$ as a true label, $Z$ as a selected label, and $X$ as an observed condition. When $n=2$, this classification is the same as the medical test in essence. Its main task is to find the best dividing point $x'$ (as shown in Fig. 4) to maximize the average log-likelihood, e.g., $-NH(Y|\theta_Z)$. Since the Shannon mutual information $I(Y; Z)$ ascertains the upper limit of the average log-likelihood, the main task is also to find the best $x'$ that maximizes $I(Y; Z)$ instead of $I(Y; X)$.

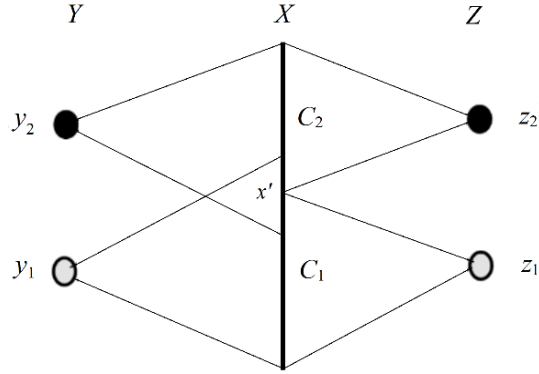

Fig. 4. Binary unseen instance classification for a given mixture model. Shannon's mutual information $I(Y; Z)$ varies with $x'$ or partition $\{C_1, C_2\}$.

For given $P(Y)$ and $P^*(X|Y)$. The semantic mutual information $I(Y; \theta_Z)$ ($\theta_Z$ is a predictive model or a fuzzy set on $V$ with respect to $Z$) is the function of partitioning $C=\{C_1, C_2, ..., C_n\}$:

$$I(Y;\theta_Z|C) = \sum_j \sum_i P(C_j) P(y_i|C_j) \log \frac{T(\theta_{zj}|y_i)}{T(\theta_{zj})} \tag{3.16}$$

where

$$P(C_j) = \sum_{x_k \in C_j} P(x_k), \ P(y_i|C_j) = \sum_{x_k \in C_j} P(y_i, x_k), \text{ and } T(\theta_{zj}) = \sum_i T(\theta_{zj}|y_i) P(y_i).$$

In the following, we introduce the CM algorithm for the best partitioning $C^*$ that maximizes the average log-likelihood and the Shannon mutual information.

A partitioning $C=\{C_1, C_2, ..., C_n\}$ of $A$ ($X \in A$) ascertains a Shannon channel:

$$P(z_j|Y) = \sum_{x_i \in C_j} P^*(x_i|Y), \ j=1,2,...,n. \tag{3.17}$$

Let the semantic channel match the Shannon channel by

$$T(\theta_{zj}|Y) = P(z_j|Y)/\max[P(z_j|Y)], \ j=1,2,...,n. \tag{3.18}$$

For given $X$ and $C$, the semantic information conveyed by $z_j$ about $Y$ is

$$I(Y;\theta_{zj}|X) = \sum_i P(y_i|X) \log \frac{T(\theta_{zj}|y_i)}{T(\theta_{zj})} . \tag{3.19}$$

We use the classifier

$$z_j = f(X) = \arg\max_{z_j} I(Y; \theta_{z_j} | X), \ j=1, 2, ..., n \tag{3.20}$$

to provides a new partitioning *C* and a new Shannon's channel, which matches the semantic channel. Repeating Eqs. (3.17)-(3.20), we can obtain the convergent *C*=*C*\*, which is the classification with maximum Shannon's mutual information $R_{max}=R(G_{max})$ (see Fig. 1). The convergence is global and can be strictly proved with the help of *R*(*G*) function (Lu, 2018a).

For example, given a binary mixture model with $P(y_1)$=0.8, $\mu_1$=30, $\mu_2$=70, $\sigma_1$=15, and $\sigma_2$=10, we use the CM algorithm for the maximum mutual information classification as follows.

Using start point *x'*=50, we have the following results. After the first iteration, x'=53. After the second iteration, *x'*=54. After the third iteration, *x'* is still 54. So, the best *x'* is *x*\*=54. The number of iterations is 3.

Even if the star point is terrible, the convergence is still fast and reliable. For example, using the start point *x'*=11, we only need two more iterations to get *x*\*=54. The number of iterations is 5.

For more details about the CM algorithm for maximum mutual information classifications, see (Lu, 2018a). This algorithm allows the source *P*(*Y*) to change. The classifier will vary with the source and hence can overcome the class-imbalance problem. This algorithm can be used to explain that the denotation of "old people" in natural language varies with the population age distribution (Lu, 2018c).

## 4. Comparing the CM-EM, EM, and MM Algorithms for Mixture Models

### *4.1 Relationships between the CM-EM, EM, and MM algorithms*

The basic formula of the EM algorithm (see Eq. (2.7)) can be simplified into

$$L_X(\theta) \geq Q+H. \tag{4.1}$$

We can express it by cross-entropies:

$$-H_\theta(X) \geq -H(X, Y|\theta) + H_\theta(Y|X). \tag{4.2}$$

Add the Shannon entropy *H*(*X*) to both sides. Then we have

$$-H(P||P_\theta) \geq H(X)-H(X,Y|\theta)+H_\theta(Y|X)=[H(X)+H_\theta(Y)-H(X,Y|\theta)]-[H_\theta(Y)-H_\theta(Y|X)]=G-R''. \tag{4.3}$$

Hence $H(P||P_\theta) \leq R''-G$, which is like the basic Eq. (3.13) in the CM-EM algorithm. Therefore, the two algorithms are interlinked. However, the CM-EM algorithm does not use Jensen's inequality.

In the M-step of the EM algorithm, maximizing *Q* is equivalent to minimizing the cross-entropies $H_\theta(Y)$ and $H(X|\theta)$. We can obtain $P(Y)=P^{+1}(Y)$ (see Eq. (3.6)) and the parameters as well as those in Eq. (3.7) for the MG-step. There is a relationship:

E-step of the EM algorithm = E1-step of the CM-EM algorithm,
M-step of the EM algorithm ≈ (E2-step + MG-step) of the CM-EM algorithm.

Maximizing *Q* is equivalent to minimizing $H_\theta(Y)$ and $H_\theta(X|Y)$. However, the M-step of the EM algorithm only modifies *P*(*Y*) one time. The E2-step modifies *P*(*Y*) many times until $P^{+1}(Y)=P(Y)$. Example 3 in Section 4.2 shows that the modification of *P*(*Y*) in the E2-step may decrease *Q* and $H(P||P_\theta)$ simultaneously.

Consider the MM algorithm (Neal and Hinton, 1999). The objective function is

$$F=Q+NH(Y)=-NH(X,Y|\theta)+NH(Y) \approx -NH(X|\theta). \tag{4.4}$$

We may treat *F* as the negative posterior cross-entropy of *X*. *F* and *G* increase or decrease simultaneously. The "≈" is used because $P^{+1}(Y) \neq P(Y)$ and hence $H_\theta(Y) \neq H(Y)$ in general. If we replace *H*(*Y*) with $H_\theta(Y)$, *F* will be equal to $-NH(X|\theta)$. Since $G=H(X)-H(X|\theta)$, maximizing *F* in the M-step of the MM algorithm is similar to maximizing *G* in the MG-step of the CM-EM algorithm. However, the E-step of the MM algorithm also maximizes *F*. The E1 and E2-steps of the CM-EM algorithm are different. They let *P*(*Y*|*X*) or *P*(*Y*) match *θ* and *P*(*X*) for the minimum of $H(P||P_\theta)$ instead of the

maximum of $G$. The reason is that $G$ might be greater than $G^*$ and therefore should be decreased (see Fig. 3).

### 4.2 Comparison of iteration numbers

The EM and improved EM algorithms generally need more than ten iterations for the convergence (Neal and Hinton, 1999; Springer and Urban, 2014; Huang and Chen, 2017) whereas the CM-EM algorithm needs less than ten iterations in most cases.

Neal and Hinton compared the iteration numbers of the EM algorithm and the MM algorithm with an example, in which a mixture component is overlapped by another. Now we use the same example to check the iteration number of using the CM-EM algorithm.

**Example 3**. Real and start parameters including mixture ratios as shown in Table 5 are obtained from the example in Neal and Hinton's paper (1999). The transform formula from the original $X'$ to $X$ that Table 5 uses is $X=20(X'-50)$. Using the CM-EM algorithm, we obtain $H(P||P_\theta)=0.00072$ bit after 9 E1 and E2-steps and 8 MG-steps.

**Table 5.** Real and guessed model parameters and iterative results of Example 3

|  | Real parameters | | | Starting parameters $H(P||P_\theta)=0.68$ bit | | | Parameters after 9 E2-steps $H(P||P_\theta)=0.00072$ bit | | |
|---|---|---|---|---|---|---|---|---|---|
|  | $\mu^*$ | $\sigma^*$ | $P^*(Y)$ | $\mu$ | $\sigma$ | $P(Y)$ | $\mu$ | $\sigma$ | $P(Y)$ |
| $y_1$ | 46 | 2 | 0.7 | 30 | 20 | 0.5 | 46.001 | 2.032 | 0.6990 |
| $y_2$ | 50 | 20 | 0.3 | 70 | 20 | 0.5 | 50.08 | 19.17 | 0.3010 |

Table 6 shows the iteration numbers and final parameters with three algorithms. When the most parameters of using the CM-EM algorithm are closer to the real parameters than those of using the EM or MM algorithm, the iteration number of using the CM-EM algorithm is about half of that of using the MM algorithm and is about a quarter of that of using standard EM algorithm.

**Table 6.** The iteration numbers and final parameters with different algorithms

| Algorithm | Sample size | Iteration number | Convergent parameters | | | | |
|---|---|---|---|---|---|---|---|
|  |  |  | $\mu_1$ | $\mu_2$ | $\sigma_1$ | $\sigma_2$ | $P(y_1)$ |
| EM | 1000 | about 36 | 46.14 | 49.68 | 1.90 | 19.18 | 0.731 |
| MM | 1000 | about 18 | 46.14 | 49.68 | 1.90 | 19.18 | 0.731 |
| CM-EM | ∞ | 9 | 46.001 | 50.08 | 2.03 | 19.17 | 0.699 |
| Real parameters |  |  | 46 | 50 | 2 | 20 | 0.7 |

The iterative process is shown in Fig. 4.

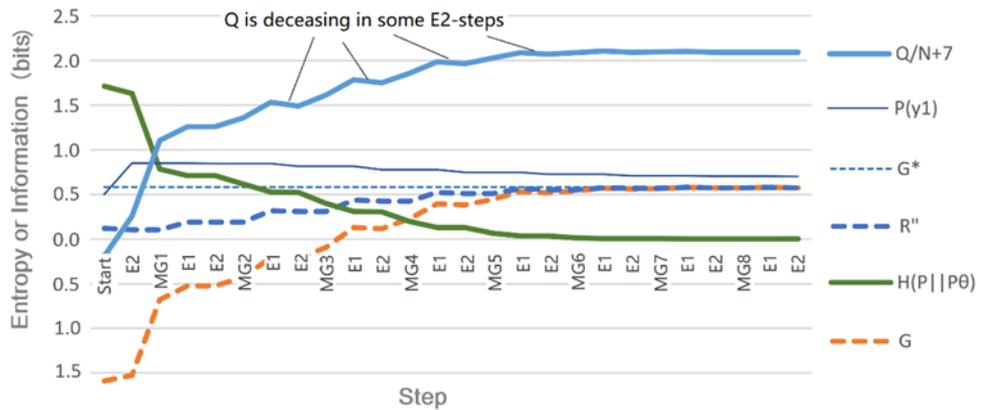

Fig. 5. The iterative process of the CM-EM algorithm for Example 3. This example was used by Neal and Hinton to compare the MM algorithm with the EM algorithm. Some E2-steps decrease *Q*. The relative entropy is less than 0.001 bit after 8 MG-steps and 9 E2-steps.

Example 3 is also a counterexample against the popular convergence proof of the EM algorithm because it also indicates that $Q$ and $L_X(\theta)$ are not positively related.

## 5. Discussion

The core concept of the CM-EM algorithm is to minimize $R$-$G$ repeatedly so that $H(P||P_\theta)$ approaches zero whereas the core concept of the EM is to maximize $Q$ repeatedly so that the log likelihood $\log L_X(\theta)$ reaches its maximum. Using $H(P||P_\theta)$ instead of $Q$, $F$, or $\log L_X(\theta)$ as the objective function, we can easily find whether the convergence is local. Exploiting the properties of $R(G)$ function, we can prove that the convergence of the CM-EM algorithm is global.

From the perspective of the semantic information theory, the EM algorithm for Mixture Models (EM-MM) is different from the EM algorithm for the Hidden Markov Models (EM-HMM), where $Y=f(X)$. In the EM-MM, we need to let Shannon's channel $P(Y|X)$ match the source $P(X)$, whereas, in the EM-HMM, we need to let the source with parameters $P_\theta(X)$ match the given Shannon's channel $P(Y|X)$ (Dempster, Laird, and Robin, 1977). Since in the EM-HMM there is no requirement for modifying $P(Y)$, the convergence of the EM-HMM has no problem. However, the convergence proof of the EM-HMM via $Q$'s increasing is nonideal. Using the semantic information method, we should be able to provide a clear convergence proof, which will be discussed elsewhere.

The CM algorithm may be used for mixture models and maximum mutual information classifications. For mixture models, it becomes the CM-EM algorithm. For maximum mutual information classifications, we need to maximize $R$ and $G$ so that $R$ and $G$ reach their maxima simultaneously. If instances are visible, we let the semantic channel and the Shannon channel mutually match only one time (Lu, 2018c). If instances are unseen, we need to let the two channels mutually match repeatedly (Lu, 2017a). It is for label learning to let the semantic channel match the Shannon channel whereas it is for label selection to let the Shannon channel match the semantic channel. The CM algorithm uses the truth function instead of the Bayesian posterior as the inference tool. It is a general method in the new mathematical framework for statistical learning (Lu, 2018b).

The semantic information method exploits sampling distributions instead of sampling sequences. With sampling distributions, we can use the cross-entropy and mutually cross-entropy more conveniently and hence deal with large samples better.

For analytical convenience, the observed data we used are only one dimensional. For multi-dimensional observed data, the computation of the CM-EM algorithm must be more complicated, but the iteration numbers should also be smaller.

The semantic information theory with $R(G)$ function basing the CM-EM algorithm was provided twenty years ago (Lu, 1993 and 1999). At that time, $R(G)$ function was used only for optimizing image compression with consideration of visual discrimination. This theory was not concerned by researchers as the author had expected. The CM algorithm for statistical learning is also a test for this theory. The author wishes that this theory could attract more researchers' attention.

## 6. Conclusions

By analyzing the EM algorithm for mixture models with the cross-entropy method, this paper revealed that 1) $Q$ may and should decrease in some E-steps; 2) the Shannon channel from the E-step is improper in general; 3) there exists a unique mixture ratio vector $P(Y)$ that matches observed data distribution $P(X)$ and parameter set $\theta$ so that the Shannon channel is qualified. So, the author proposed the CM-EM algorithm, an improved EM or MM algorithm. The CM-EM algorithm adds a step for the unique $P(Y)$ into the E-step and only optimizes $\theta$ to maximize the semantic mutual information $G$ in the M-step, e. g., the MG-step. To prove the convergence of the CM-EM algorithm, this paper first proved that the minimum relative entropy is equal to the minimum *R-G* (*R* is the Shannon mutual information). Then it used the variational method and iterative method that Shannon et al. used for analyzing the rate-distortion function to prove that every step of the CM-EM algorithm reduces the relative entropy $H(P||P_\theta)$ until $R=G$ and $H(P||P_\theta)=0$. Making use of the properties of $R(G)$ function the author proposed before (Lu, 1999), this paper proved that the convergence is global. The paper provided three different examples of mixture models, for which the CM-EM algorithm needs only 5, 5, and 9 iterations respectively. For the last example, the EM algorithm needs about 36 iterations, and the MM algorithm proposed by Neal and Hinton needs about 18 iterations. Theoretical analyses and computational experiments indicated that the CM-EM algorithm is fast and reliable. For using optimized mixture models, the paper introduced the CM algorithm for maximum mutual information classifications and showed that its iteration is also very fast.

## Appendix

*Appendix I: Optimizing P(Y|X) and P(Y) for the minimum of R-G*

To optimize *P(Y|X)*, Order

$$\frac{\partial F}{\partial P(y_j|x_i)} = \frac{\partial}{\partial P(y_j|x_i)} \{\sum_j\sum_i P(x_i)P(y_j|x_i)\log\frac{P(y_j|x_i)}{P(y_j)} \\ -\sum_j\sum_i P(x_i)P(y_j|x_i)\log\frac{P(x_i|\theta_j)}{P(x_i)} + \mu_i\sum_j P(y_j|x_i) + \alpha\sum_j P(y_j)\} = 0. \quad (1)$$

Hence

$$P(x_i)[1+\log P(y_j|x_i)] - P(x_i)\log[P(x_i|\theta_j)/P(x_i)] + \mu_i = 0, \\ \log[P(y_j|x_i)/P(y_j)] = \log[P(x_i|\theta_j)/P(x_i)] - (\mu_i+1)/P(x_i). \quad (2)$$

Order $\log\lambda_i=(\mu_i+1)/P(x_i)$, we have

$$P(y_j|x_i) = P(y_j)P(x_i|\theta_j)/\lambda_i, \ i=1, 2, ..., n; \ j=1, 2. \quad (3)$$

Since the second order partial derivative is greater than 0, this *P(Y|X)* minimizes *I(X;Y)-I(X;θ)*. Since $P(Y|x_i)$ is normalized, the optimized *P(Y|X)* is

$$P^*(y_j|x_i) = P(y_j)P(x_i|\theta_j)/\sum_k P(y_k)P(x_i|\theta_k), \ i=1,2,...,n; \ j=1,2. \quad (4)$$

To optimize *P(Y)*, order

$$\frac{\partial}{\partial P(y_j)}[\sum_j\sum_i P(x_i)P(y_j|x_i)\log\frac{P(y_j|x_i)}{P(y_j)} + \mu_i\sum_j P(y_j|x_i) + \alpha\sum_j P(y_j)] = 0. \quad (5)$$

Hence

$$-\sum_i P(x_i)P(y_j|x_i)/P(y_j) + \alpha = 0, \quad (6)$$

$$P(y_j) = \frac{1}{\alpha}\sum_i P(x_i)P(y_j|x_i). \quad (7)$$

Since the second order partial derivative is greater than 0, this *P(Y)* minimizes *I(X;Y)-I(X;θ)*. Since $\sum_j P^*(y_j)=1$, $\alpha=1$. Therefore, the optimized *P(Y)* is

$$P^*(y_j) = \sum_i P(x_i)P(y_j|x_i), j=1, 2, ..., n. \quad (8)$$

**QED.**